\documentclass[final,times,5p,twocolumn]{elsarticle}

\usepackage{lineno}
\usepackage{amssymb}
\usepackage{amsmath}
\usepackage[ruled,linesnumbered]{algorithm2e}
\usepackage{xcolor}
\usepackage{mdframed}
\usepackage[ruled,linesnumbered]{algorithm2e}
\usepackage{amsmath,amssymb,amsfonts}
\usepackage{graphicx}
\usepackage{epstopdf}
\usepackage{subfig}
\usepackage{gensymb}
\usepackage{multirow,booktabs,color,soul,threeparttable}
\usepackage{xcolor}
\usepackage{mdframed}
\usepackage{hyperref}
\usepackage{cleveref}
\usepackage{rotating}
\usepackage[most]{tcolorbox}
\usepackage{tabularx}
\usepackage{array}

\usepackage{natbib}

\newcommand{\RNum}[1]{\uppercase\expandafter{\romannumeral #1\relax}}
\definecolor{hl}{rgb}{0.75,0.75,0.75}
\sethlcolor{hl}
\definecolor{emph}{rgb}{0,0,1}

\begin{document}

\begin{frontmatter}

\title{AutoSynth: Automated Workflow Optimization for High-Quality Synthetic Dataset Generation via Monte Carlo Tree Search}

\author[sii]{Shuzhen Bi} \ead{SA22916003@mail.ustc.edu.cn}
\author[aiedu,ecnu]{Chang Song} \ead{19015347605@163.com}
\author[aiedu,ecnu]{Siyu Song} \ead{siyusong00@gmail.com}
\author[ecnu]{Jinze Lv} \ead{51275901080@stu.ecnu.edu.cn}
\author[aiedu]{Jian Chen} \ead{52295901041@stu.ecnu.edu.cn}
\author[aiedu,ecnuedu]{Xinyun Wang} \ead{51264108008@stu.ecnu.edu.cn}
\author[sii,aiedu,ecnu]{Aimin Zhou} \ead{amzhou@cs.ecnu.edu.cn}
\author[aiedu]{Hao Hao \corref{mycorrespondingauthor}}
\ead{hhao@mail.ecnu.edu.cn}\cortext[mycorrespondingauthor]{Corresponding author: Hao Hao}

\affiliation[sii]{organization={Shanghai Innavation Institute}, city={Shanghai}, postcode={200231}, country={China}}
\affiliation[aiedu]{organization={Shanghai Institute of AI for Education},%Department and Organization
addressline={East China Normal University}, 
city={Shanghai},
postcode={200062}, 
country={China}}
\affiliation[ecnuedu]{organization={Department of Education Information Technology},addressline={East China Normal University}, city={Shanghai}, postcode={200062}, country={China}}
\affiliation[ecnu]{organization={School of Computer Science and Technology},%Department and Organization
addressline={East China Normal University}, 
city={Shanghai},
postcode={200062}, 
country={China}}

% Abstract
\begin{abstract}
Supervised fine-tuning (SFT) of large language models (LLMs) for specialized tasks requires high-quality training datasets, yet manual curation remains prohibitively expensive and time-consuming. While synthetic data generation offers a scalable alternative, its effectiveness depends on carefully designed generation workflows---complex, multi-stage processes integrating prompt engineering, programmatic logic, and model orchestration. Existing automated workflow optimization methods fail to address the cold start problem, as they require labeled datasets for reward modeling, particularly problematic for subjective, open-ended tasks where objective ground truth is unavailable. We introduce AutoSynth, a framework that automates workflow discovery and optimization without reference datasets by reframing the problem as a Monte Carlo Tree Search guided by a novel dataset-free hybrid reward signal. This signal enables meta-learning through two components: an LLM-as-judge evaluating sample quality against dynamically regenerated task-specific metrics, and another LLM-as-judge assessing workflow code and prompt quality. Experiments on subjective educational tasks reveal that while expert-designed workflows achieve higher human preference rates (96-99\% win rates compared to AutoSynth's 40-51\%), models trained on AutoSynth-generated data dramatically outperform baseline models (40-51\% vs. 2-5\% win rates) and surpass expert-designed workflows on certain metrics, suggesting the discovery of quality dimensions not fully captured by human evaluators. These results are achieved while reducing human effort from 5-7 hours to just 30 minutes (>90\% reduction). AutoSynth addresses the critical cold start challenge in data-centric AI, providing a scalable, cost-effective method for developing specialized LLMs for subjective tasks lacking objective ground truth. The code is available at \url{https://github.com/bisz9918-maker/AutoSynth}.
\end{abstract}

%% Keywords

\end{frontmatter}

% \linenumbers

\section{Introduction}

The rapid advancement of large language models (LLMs) has ushered in a new era of artificial intelligence, where model specialization through supervised fine-tuning (SFT) is the dominant paradigm for adapting general-purpose models to domain-specific applications~\cite{ouyang2022training, zhou2023lima, anisuzzaman2024fine}. However, SFT fundamentally depends on high-quality, task-aligned training data~\cite{zhou2024datacentricAI, li2025data}. Manually curating such datasets requires substantial domain expertise, extensive time, and significant financial resources, creating a critical bottleneck in the democratization of specialized AI~\cite{levin2024performance, jimenez2023swe}.

Synthetic data generation, leveraging LLMs to create training examples, has emerged as a promising solution to this data scarcity challenge~\cite{wang2022self, honovich2022unnatural, ye2022zerogen}. Yet, the quality of synthetic data critically depends on the design of the \textit{generation workflow}---a complex, multi-stage process integrating prompt engineering, programmatic logic, and model orchestration. A poorly designed workflow can yield data lacking diversity~\cite{chen2024diversity}, containing factual errors, or failing to align with nuanced task requirements~\cite{long2024llms}.

Recent work has begun to automate workflow design. Frameworks like AFlow~\cite{zhang2024aflow} and AutoFlow~\cite{li2024autoflow} reformulate this as a search problem, using MCTS or reinforcement learning to refine implementations. While effective for well-defined tasks, they share a fundamental limitation: their optimization presupposes a pre-existing labeled dataset to provide a reward signal (e.g., measuring correctness against a held-out answer key). This dependency renders them inapplicable to the most common and critical scenario: creating datasets for novel tasks where no reference data or ground truth exists.

This cold start problem is particularly severe for subjective, open-ended tasks (e.g., educational content, creative writing, personalized tutoring), where: (1) ``correct" answers are ill-defined, (2) quality is multi-dimensional and context-dependent, and (3) no validation benchmarks exist. For instance, generating high-quality math explanations requires pedagogical soundness, conceptual accuracy, and student engagement---criteria that defy simple correctness metrics~\cite{lu2025fine}. Existing automation methods offer no solution, forcing practitioners back to costly, non-scalable manual workflow engineering.

We address this critical gap by introducing \textbf{AutoSynth}, the first framework to automate synthetic data workflow optimization \textit{without requiring any reference datasets}. Our key insight is to replace dataset-dependent rewards with a novel \textit{dataset-free hybrid reward signal} computed entirely by LLMs. This signal guides a Monte Carlo Tree Search (MCTS) and uniquely enables a meta-learning process by combining two LLM-as-judge components:

\begin{enumerate}
    \item \textbf{Sample Quality Assessment}: An evaluator LLM assesses generated samples against \textit{task-specific quality metrics that are themselves dynamically regenerated and refined} at each iteration, allowing the system's understanding of ``quality" to co-evolve with its generation capabilities.
    
    \item \textbf{Workflow Quality Assessment}: An optimizer LLM evaluates the workflow's own code structure, prompt design, and logical coherence, preventing optimization from converging on brittle or unmaintainable implementations.
\end{enumerate}

This process requires only a brief task description and minimal human alignment (approx. 30 minutes) to establish the initial baseline, after which the MCTS optimization proceeds fully autonomously.

We validate AutoSynth on two challenging subjective tasks in the educational domain: \textit{math concept explanation} and \textit{interdisciplinary lesson planning}. Our experiments reveal a nuanced and critical finding:
\begin{itemize}
    \item While expert-designed workflows still achieve near-perfect human preference rates (96--99\%), models trained on AutoSynth data (40--51\% win rates) dramatically outperform the baseline model (2--5\% win rates).
    
    \item More importantly, AutoSynth-trained models consistently surpassed expert-designed workflows on metrics, suggesting the discovery and optimization of quality dimensions not fully captured by static human evaluation.
    
    \item This high-quality result is achieved while reducing human effort by over 90\% (from 5--7 hours for expert design to 30 minutes for AutoSynth initialization).
    
    \item Ablation studies confirm that all components of our design---the hybrid reward, human-in-the-loop initialization, and iterative metric regeneration---are critical to system performance.
\end{itemize}

Our contributions are as follows:
\begin{enumerate}
    \item We propose the workflow optimization framework for synthetic data that operates without reference datasets, addressing the cold start problem for subjective tasks.
    \item We introduce a dataset-free hybrid reward signal that synergistically combines sample-level and workflow-level quality assessment.
    \item We develop a dynamic metric regeneration strategy that enables a meta-learning process, where evaluation criteria co-evolve with generation capabilities.
    \item We provide strong empirical evidence that this automated framework, while reducing human effort by over 90\%, generates high-quality data that dramatically outperforms baselines and achieves superior performance metrics.
\end{enumerate}

This work represents a significant step toward fully automated, data-centric AI development, particularly for the vast landscape of subjective tasks where traditional supervised learning fails due to the absence of objective ground truth.

\section{Related Work}

Our work bridges two previously disparate research streams: automated workflow optimization for LLM-based systems and synthetic data generation for supervised fine-tuning. We review both domains and articulate how AutoSynth addresses their fundamental incompatibility.

\subsection{Automated Workflow Optimization for LLMs}

Recent advances in agentic AI have motivated frameworks that automatically construct and optimize multi-step LLM-based workflows. We focus on systems that treat workflow design as a search or optimization problem.

\textbf{Search-Based Workflow Optimization.} AFlow~\cite{zhang2024aflow} pioneered the formulation of workflow optimization as a Monte Carlo Tree Search (MCTS) problem over code-represented execution graphs. Workflows are modeled as directed acyclic graphs where nodes encapsulate LLM calls or computational operations, and edges encode control flow and data dependencies. AFlow iteratively refines these structures through tree-structured experience accumulation and execution-based feedback, demonstrating substantial performance gains across six benchmark datasets. AutoFlow~\cite{li2024autoflow} extends this paradigm by incorporating reinforcement learning to guide the search process, while FlowMind~\cite{zeng2023flowmind} explores automated prompt composition within workflow contexts. These approaches share a common architectural principle: they decompose complex tasks into modular, executable components and leverage search algorithms to discover optimal configurations.

\textbf{The Evaluation Bottleneck.} Despite their sophistication, all existing workflow optimization frameworks share a critical limitation: they require \textit{pre-existing labeled validation datasets} to compute reward signals. AFlow, for instance, evaluates candidate workflows by executing them on held-out test cases and measuring task-specific success metrics (e.g., correctness rates for mathematical problem-solving). This dependency renders these systems inapplicable to the most prevalent real-world scenario---creating datasets for novel tasks where no ground-truth evaluation data exists. Our work directly addresses this fundamental constraint.

\textbf{Connections to AutoML.} These agentic workflow optimization methods represent a natural evolution of automated machine learning (AutoML)~\cite{he2021automl}. While classical AutoML focuses on low-level decisions such as hyperparameter tuning and architecture search, LLM-driven workflow automation~\cite{jin2025llm} operates at a higher level of abstraction, automating prompt engineering, tool orchestration, and multi-step reasoning chains. AutoSynth pushes this frontier further by automating not just the workflow itself, but also the evaluation criteria that guide its optimization.

\subsection{Synthetic Data Generation for Supervised Fine-Tuning}

The success of supervised fine-tuning (SFT) for domain adaptation hinges critically on training data quality~\cite{zhou2023lima}. Synthetic data generation has emerged as a scalable alternative to human annotation, yet ensuring quality remains a persistent challenge.

\textbf{Generation Paradigms.} Self-instruct~\cite{wang2022self} established the foundational paradigm of bootstrapping instruction-following datasets by prompting LLMs with seed examples. Distillation approaches~\cite{gu2023minillm} leverage powerful teacher models to generate training data for smaller student models, effectively transferring capabilities across model scales. While these methods reduce annotation costs, they inherit biases from the generating models and often produce data with limited diversity or factual inaccuracies~\cite{ji2023survey}.

\textbf{Quality Enhancement Techniques.} More sophisticated approaches incorporate iterative refinement mechanisms. Self-critique methods~\cite{madaan2023self} enable models to improve their outputs through self-generated feedback loops, while verification-based techniques~\cite{chen2023teaching} employ external validators (e.g., unit tests for code generation) to filter low-quality samples. Preference-based approaches~\cite{ouyang2022training} use reward models to score and select high-quality generations. However, these techniques operate within \textit{static, manually designed generation pipelines}, limiting their adaptability to diverse task requirements.

\textbf{Persistent Quality Challenges.} The literature identifies four fundamental challenges in synthetic data generation that our framework explicitly targets: (1) \textit{Faithfulness}—--mitigating hallucinations and factual errors~\cite{ji2023survey}; (2) \textit{Diversity}—--preventing mode collapse in generation~\cite{chang2024quality}; (3) \textit{Subjectivity}—--handling tasks with ill-defined correctness criteria~\cite{yu2023metamath}; and (4) \textit{Controllability}—--precisely steering generation to meet nuanced specifications~\cite{liang2024controllable}. Existing methods address these challenges in isolation through task-specific heuristics, lacking a unified optimization framework.

\subsection{Bridging the Gap: From Parallel Development to Convergence}

The domains of workflow optimization and synthetic data generation have evolved largely independently, creating a critical gap. Workflow optimization systems (e.g., AFlow) excel at \textit{process optimization}---given a fixed evaluation metric, they find efficient execution strategies. However, they cannot define or refine the metric itself. Conversely, synthetic data research focuses on \textit{output quality}—--developing techniques like self-critique and preference learning to improve individual samples—--but relies on static generation pipelines designed through manual trial-and-error.

AutoSynth represents the first framework to  bridge this gap. By introducing a dataset-free hybrid reward signal that evaluates both workflow structure and sample quality, we enable simultaneous optimization of the generation process and the quality criteria themselves. This creates a meta-learning system where improved workflows produce better data, which in turn informs more sophisticated evaluation metrics, which then guide further workflow refinement. This synergistic approach fundamentally addresses the cold start problem: in the absence of ground-truth data, the system bootstraps both its generation capabilities and its understanding of quality from minimal human guidance. This synergistic approach fundamentally solves the cold start problem. In the absence of ground-truth data, AutoSynth bootstraps both its generation capabilities and its understanding of quality from minimal human guidance, enabling automated dataset creation in subjective domains where it was previously infeasible.

\section{The AutoSynth Framework}
\label{sec:meta_learning}
\begin{figure*}[ht!]
    \centering
    \includegraphics[width=0.95\textwidth]{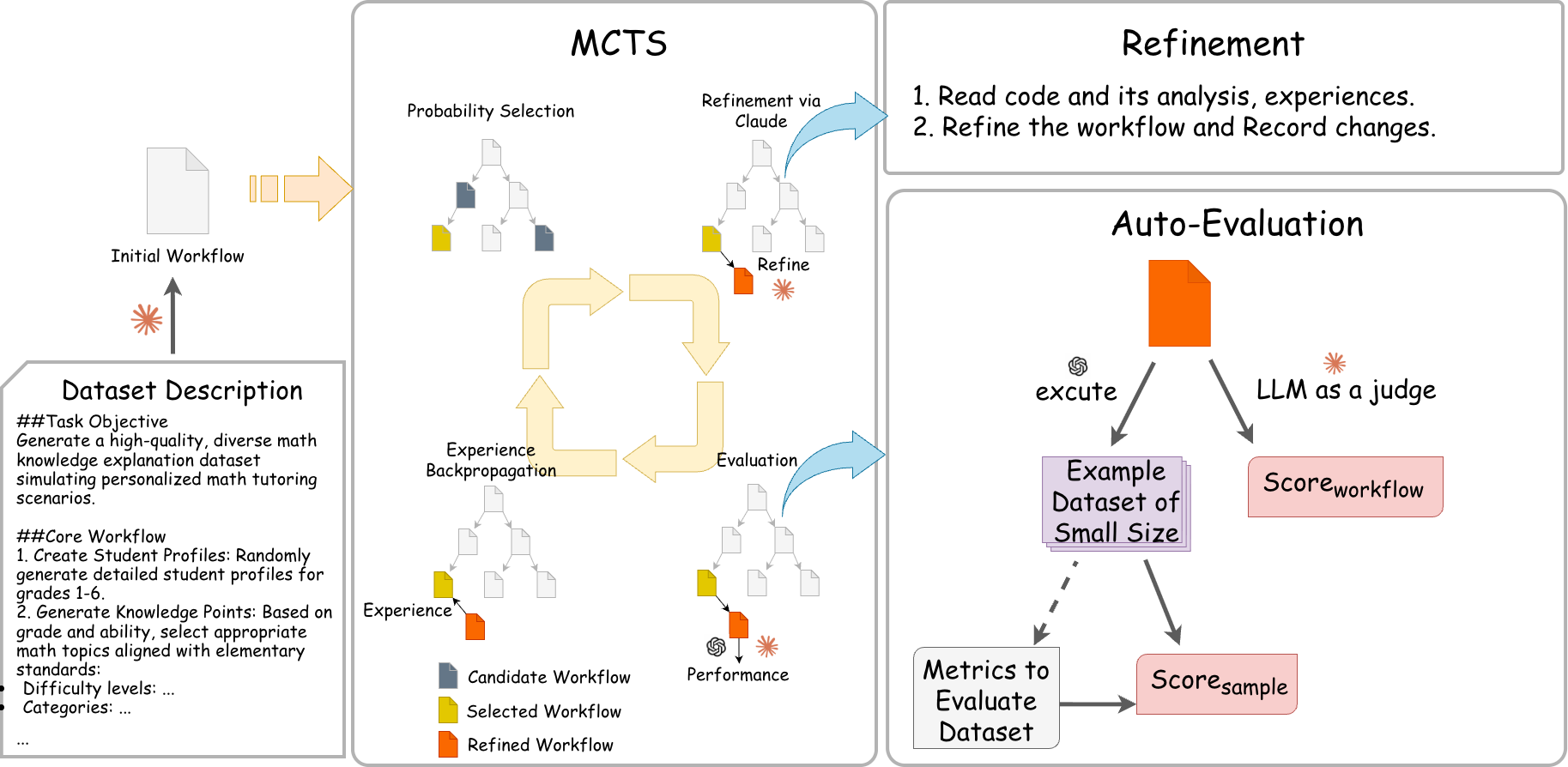}
    \caption{Overview of the AutoSynth framework. The system operates in two phases: \textbf{(Left)} Human-in-the-loop initialization generates a functional baseline workflow $W_0$ through 1-3 iterations of LLM generation, execution, and human feedback; \textbf{(Middle \& Right)} MCTS-driven optimization iteratively refines workflows through selection, refinement, evaluation, and backpropagation, guided by a hybrid reward signal combining sample quality ($Score_{sample}$) and workflow quality ($Score_{workflow}$).}
    \label{fig:framework}
\end{figure*}

This section provides a detailed, step-by-step description of our methodology, with an emphasis on reproducibility. Figure~\ref{fig:framework} presents a comprehensive architectural overview of the AutoSynth system, illustrating the three core phases: human-in-the-loop initialization, MCTS-driven workflow optimization, and the dynamic hybrid reward signal.

\subsection{Problem Formulation: Synthesis as Dataset-Free Search}

We formalize the automated dataset generation problem as follows: Given a natural language description $\mathcal{T}$ of a desired dataset, our objective is to discover an optimal workflow $W^*$ that generates a dataset $D$ maximizing a quality function $Q(D)$:
$$W^* = \arg\max_{W \in \mathcal{W}} Q(D_W)$$
where $\mathcal{W}$ denotes the space of all possible workflows and $D_W$ is the dataset produced by workflow $W$.

The fundamental challenge is that $Q$ is neither predefined nor computable via comparison against reference data, as no ground-truth dataset exists. We address this by reformulating dataset synthesis as a search problem over the space of executable workflows, where the quality function $Q$ is itself dynamically constructed through LLM-based evaluation.

We employ Monte Carlo Tree Search (MCTS) to construct a search tree $\mathcal{T} = (\mathcal{V}, \mathcal{E})$ where each node $v \in \mathcal{V}$ represents a complete workflow implementation, and each directed edge $e \in \mathcal{E}$ represents a modification (mutation) applied to transform a parent workflow into a child workflow. The search is guided by a novel dataset-free hybrid reward signal, detailed in Section~\ref{sec:reward}.

\subsection{Workflow Representation and Initialization}

\textbf{Workflow Structure.} A workflow $W$ is formally defined as a composite entity $W = \langle P, C \rangle$, where:
\begin{itemize}
    \item $P$ represents structured prompts, including system instructions, few-shot exemplars, and generation templates that guide LLM behavior.
    \item $C$ represents executable Python code implementing control logic, data transformations, error handling, and orchestration of multiple LLM API calls.
\end{itemize}
This dual representation combines the flexibility of natural language steering (via prompts) with the precision of programmatic control (via code).

\textbf{Human-in-the-Loop Initialization.} AutoSynth begins with a brief human-guided phase to align the search with user intent and establish a functional baseline.
\begin{enumerate}
    \item An \textit{Optimizer LLM} (Claude Sonnet 3.7) receives the task description $\mathcal{T}$ and generates an initial workflow $W_{\text{init}}$.
    \item $W_{\text{init}}$ is executed to produce a small batch of sample data (5-10 instances).
    \item A human expert reviews the workflow (code and prompts) and the samples, providing concise feedback on correctness, alignment, and potential improvements.
    \item The Optimizer LLM incorporates this feedback to produce a revised workflow $W'_{\text{init}}$.
    \item Steps 2-4 repeat for 1-3 iterations until a functional baseline $W_0$ is obtained that successfully executes and produces reasonable data.
\end{enumerate}
This process ensures the search begins in a promising region of the workflow space and captures nuanced user intentions. The resulting $W_0$ becomes the root node of the MCTS search tree.

\subsection{MCTS-Driven Workflow Optimization}

The MCTS optimization loop (Figure~\ref{fig:framework}, middle) iteratively refines the baseline workflow by balancing exploration and exploitation.

\subsubsection{Selection}
The search begins by selecting a promising node. We identify the top-3 workflows in the tree ranked by their hybrid reward scores $R(W)$ (defined in Section~\ref{sec:reward}). We then compute a probability distribution over these candidates:
$$p_i = \frac{R(W_i)}{\sum_{j=1}^{3} R(W_j)}$$
and sample one workflow $W_{\text{selected}}$ according to this distribution. This probabilistic selection balances exploitation of high-scoring workflows with exploration of other promising candidates.

\subsubsection{Refinement}
The selected workflow $W_{\text{selected}}$ is provided to the Optimizer LLM (Claude Sonnet 3.7), which acts as an expert programmer and prompt engineer. The LLM receives:
\begin{itemize}
    \item The original task description $\mathcal{T}$.
    \item The complete code and prompts of $W_{\text{selected}}$.
    \item All accumulated experiences from child workflows (modification descriptions, reward scores, and qualitative feedback).
    \item Actionable improvement suggestions for $W_{\text{selected}}$.
\end{itemize}
The Optimizer proposes a targeted modification (e.g., code changes, prompt enhancements, structural adjustments) to produce a new child node $W_{\text{new}}$. The description of this modification (e.g., Added self-verification step for mathematical correctness) is recorded as an \textit{experience} annotation on the edge connecting $W_{\text{selected}}$ to $W_{\text{new}}$.

\subsubsection{Evaluation}
The newly generated workflow $W_{\text{new}}$ is executed to produce a small evaluation batch of $n=5\text{-}10$ data samples. This batch serves as the input for our Dataset-Free Hybrid Reward Signal computation (detailed in Section~\ref{sec:reward}), which concurrently assesses both the \textbf{quality of the generated samples} and the \textbf{structural integrity of the workflow}.

\subsubsection{Backpropagation}
The hybrid reward $R(W_{\text{new}})$ computed for the new workflow is recorded and associated with the $W_{\text{new}}$ node. Unlike traditional MCTS, which propagates averaged scores, AutoSynth preserves the precise score for each workflow.

What is propagated to ancestor nodes is the \textit{experience record}: the modification description, the resulting reward $R(W_{\text{new}})$, and the qualitative feedback. This experience is appended to $W_{\text{selected}}$ and its ancestors, enriching the context for future refinement decisions and allowing the Optimizer LLM to learn from both successful and unsuccessful modifications.

\textbf{Termination.} The MCTS loop continues until convergence (scores stabilize) or a maximum of 30 iterations is reached. The highest-scoring workflow $W^*$ is then selected.

\subsection{Dataset-Free Hybrid Reward Signal}
\label{sec:reward}

The core innovation enabling AutoSynth to operate without reference datasets is our hybrid reward mechanism. It combines assessments of both the generated data (the product) and the workflow itself (the process). The final reward is an unweighted average of these two components:
$$R(W) = 0.5 \cdot Score_{sample}(W) + 0.5 \cdot Score_{workflow}(W)$$
This balanced formulation prevents pathological solutions, such as a brittle workflow that produces good samples but is unmaintainable, or an elegant workflow that produces mediocre data.

\subsubsection{Component 1: Dynamic Sample Quality Score ($Score_{sample}$)}
\label{sec:sample_quality}

This component assesses the quality of the workflow's output $D_W$. Its calculation is founded on a dynamic metric generation process that enables the system to co-evolve its generation and evaluation capabilities.

\textbf{Dynamic Metric Generation and Co-evolution.}
Instead of relying on a fixed, predefined metric, $Score_{sample}$ is computed against criteria that are dynamically generated at each MCTS iteration. This mechanism transforms the optimization into a meta-learning process, addressing the ``Goodhart's Law" problem: when a measure becomes a target, it ceases to be a good measure.

This co-evolutionary dynamic proceeds as follows:
\begin{enumerate}
    \item \textbf{Metric Proposal:} Given the batch of samples $D_{W_t}$ from the current workflow $W_t$, an \textit{Evaluator LLM} (GPT-5) is prompted to act as a domain expert. It proposes a comprehensive set of evaluation metrics $M_t = \{m_1, \ldots, m_k\}$ tailored to the task description $\mathcal{T}$.
    \item \textbf{Metric Structure:} Each metric $m_i$ includes a name, a detailed definition, a positive exemplar, and a negative exemplar to anchor scoring decisions.
    \item \textbf{Self-Consistency Ensemble:} To ensure robustness, the Evaluator LLM independently generates three candidate metric sets. The set with the highest average semantic overlap with the other two is selected, filtering out idiosyncratic metrics.
\end{enumerate}
Critically, this process is re-executed at every MCTS iteration. This dynamic regeneration compels the system to continuously refine its own conception of quality. As workflows improve (e.g., from generating factually correct text to generating text with clear analogies), the metrics evolve to capture these new dimensions of quality (e.g., adding an ``Analogy Clarity" metric). The system does not merely optimize toward a static goal; it refines the goal itself as it learns.

\textbf{Sample-by-Sample Evaluation.}
Given the newly generated dynamic metrics $M_t$, the Evaluator LLM (GPT-5) scores each sample $d_i \in D_{W_{\text{new}}}$ against each metric $m_j$. For each pair, the LLM provides:
\begin{itemize}
    \item A numerical score $s_{ij} \in [1, 5]$ based on the metric's exemplars.
    \item A brief qualitative justification for the score.
\end{itemize}

\textbf{Aggregation.}
The sample quality score is computed as the average across all samples and all metrics:
$$Score_{sample}(W) = \frac{1}{|D_W| \cdot |M_t|} \sum_{i=1}^{|D_W|} \sum_{j=1}^{|M_t|} s_{ij}$$
The qualitative justifications are concatenated and stored as improvement suggestions for the next MCTS refinement step.

\subsubsection{Component 2: Workflow Quality Score ($Score_{workflow}$)}
In parallel, the Optimizer LLM that performed the refinement step also conducts an introspective evaluation of the new workflow $W_{\text{new}}$ itself, assessing the \textit{process} rather than the \textit{product}. This score is based on:
\begin{itemize}
    \item \textbf{Code Quality} ($Q_{\text{code}}$): Assessed on clarity (readability, documentation), efficiency (complexity, API optimization), and robustness (error handling, edge cases).
    \item \textbf{Prompt Quality} ($Q_{\text{prompt}}$): Assessed on clarity (unambiguous instructions), specificity (detailed requirements), and effectiveness~(likelihood of eliciting desired behavior).
\end{itemize}
This component acts as a regularizer, ensuring that optimized workflows are not only effective but also generalizable, maintainable, and human-interpretable.

\section{Experiments}
\subsection{Experimental Setup}

\textbf{Tasks and Domains.} We evaluate AutoSynth on two challenging educational tasks where quality is inherently subjective and no ground-truth answers exist:

\begin{enumerate}
    \item \textbf{Math Concept Explanation}: Generating pedagogically sound explanations of mathematical concepts tailored to students at specific learning stages. 

    \item \textbf{Interdisciplinary Lesson Plan Generation}: Creating comprehensive lesson plans that integrate multiple subject areas (e.g., mathematics and history, science and literature) with authentic contexts and coherent learning objectives.
\end{enumerate}

These domains exemplify scenarios where AutoSynth's dataset-free approach provides maximum value: traditional supervised learning struggles due to the absence of objective correctness criteria, and manually designing effective generation workflows demands substantial pedagogical expertise.

\textbf{Implementation Details.} We employ a division of labor between two state-of-the-art LLMs to leverage their respective strengths:

\begin{itemize}
    \item \textbf{Optimizer LLM (Claude Sonnet 3.7)}: Responsible for workflow generation and refinement during MCTS expansion, as well as introspective evaluation of workflow code and prompt quality ($Score_{workflow}$). Claude's strong capabilities in code generation and structured reasoning make it well-suited for programmatic workflow optimization.
    
    \item \textbf{Evaluator LLM (GPT-5)}: Handles all data-centric operations including (1) executing workflows to generate training samples, (2) dynamically generating task-specific evaluation metrics from generated samples, and (3) scoring individual samples against these metrics ($Score_{sample}$). GPT-5's advanced reasoning and nuanced assessment capabilities enable reliable quality evaluation in subjective educational domains.
\end{itemize}
The MCTS optimization runs for a maximum of 30 iterations or until convergence, defined as the top-3 workflow scores stabilizing within threshold $\epsilon = 0.05$. In practice, most tasks converge within 15-25 iterations.

\textbf{Baselines and Comparisons.} We compare AutoSynth against two critical baselines:

\begin{enumerate}
    \item \textbf{Expert-Designed Workflow}: A senior educational technology researcher manually designs and iteratively refines a data generation workflow for each task. This process involves: (a) analyzing task requirements, (b) designing initial prompt templates and code structure, (c) generating sample data and manually evaluating quality, (d) iteratively refining the workflow based on observed weaknesses. This represents the current best practice for synthetic dataset creation and requires 5-7 hours of expert effort per task.
    
    \item \textbf{Base Model (Zero-Shot)}: The unmodified Qwen-Instruct-32B model evaluated directly on test queries without any task-specific fine-tuning.
\end{enumerate}

\textbf{Dataset Generation and Fine-Tuning.} For each task and method (AutoSynth, Expert-Designed, Base Model), we generate 1,000 training examples using the respective workflows (no data generation for Base Model). We then perform supervised fine-tuning (SFT) on Qwen-Instruct-32B using standard hyperparameters: learning rate $2 \times 10^{-5}$, batch size 16, 3 epochs with linear warmup and cosine decay. The Base Model serves as the pre-training checkpoint.

\textbf{Evaluation Framework.} We employ the ELMES (Evaluating Large Language Models in Educational Scenarios) framework, an open-source automated evaluation system specifically designed for assessing LLMs in educational settings that incorporates a hybrid evaluation engine that objectively quantifies traditionally subjective pedagogical metrics using an LLM-as-a-Judge methodology~\cite{wei2025elmes}. 
For all evaluations, we construct a held-out test set of 10 queries per task, ensuring no overlap with training data. 

\subsection{Main Results}

Table~\ref{tab:main_results} presents the comprehensive comparison across both educational tasks. The results demonstrate that AutoSynth successfully addresses the cold-start problem in dataset creation, achieving performance competitive with expert-designed workflows while drastically reducing human effort.

\begin{table*}[ht!]
\centering
\caption{Comparison of AutoSynth against expert-designed workflows and baseline model across two educational tasks. Human evaluation win rates represent the percentage of pairwise comparisons where each method is preferred over all other methods. Metric scores are task-specific evaluation scores (higher is better). Human effort quantifies the time investment required.}
\label{tab:main_results}
\begin{tabular}{llccc}
\toprule
\textbf{Task} & \textbf{Method} & \textbf{Human Win Rate (\%)} & \textbf{Metric Score} & \textbf{Human Effort} \\
\midrule
\multirow{3}{*}{\textbf{Math Concept Explanation}} 
& AutoSynth & 40.50 & 4.25 & 30 min \\
& Expert-Designed & 98.50 & 4.18 & $\sim$5 hours \\
& Base Model & 4.75 & 3.67 & 0 \\
\midrule
\multirow{3}{*}{\textbf{Interdisciplinary Lesson Plan}} 
& AutoSynth & 51.13 & 4.82 & 30 min \\
& Expert-Designed & 95.50 & 4.75 & $\sim$7 hours \\
& Base Model & 1.99 & 3.75 & 0 \\
\bottomrule
\end{tabular}
\end{table*}

\textbf{Key Findings:}

\begin{enumerate}
    \item \textbf{Substantial Improvement Over Base Model with Minimal Human Effort}: AutoSynth achieves 40-51\% win rates in human pairwise evaluations across all methods, dramatically outperforming the base model (2-5\% win rates) and demonstrating the critical value of task-specific dataset generation. Remarkably, this performance is obtained with only 30 minutes of human involvement for initialization and metric validation, compared to 5-7 hours required for expert workflow engineering—representing over 90\% reduction in human effort. While expert-designed workflows achieve the highest win rates (96-99\%), AutoSynth provides a compelling cost-quality trade-off for rapid deployment in resource-constrained scenarios.
    
    \item \textbf{Superior Metric Performance with Divergent Human Preferences}: AutoSynth achieves higher metric scores than expert-designed workflows across both tasks (4.25 vs 4.18 for Math Concepts, 4.82 vs 4.75 for Lesson Plans), revealing an intriguing divergence between metrical and human evaluation. While human evaluators strongly prefer expert-designed outputs in pairwise comparisons (96-99\% win rates), AutoSynth-trained models demonstrate superior performance on the formal evaluation metrics. This discrepancy suggests that expert-designed data may incorporate subtle stylistic preferences, rhetorical conventions, or pedagogical nuances that resonate strongly with human judges but are not explicitly captured by the metric definitions. Conversely, AutoSynth's optimization process may more directly target the quantifiable dimensions specified in the evaluation rubrics. Both fine-tuned approaches demonstrate substantial improvements over the base model (0.51-1.07 point gains), validating that specialized dataset creation, whether automated or manual, is essential for educational applications.
    
    \item \textbf{Bridging the Automation Gap}: The performance gap between AutoSynth (40-51\% win rate) and expert-designed workflows (96-99\% win rate) reveals both progress and remaining challenges in fully automated dataset generation. AutoSynth successfully automates the workflow design process that previously required hours of expert effort, but does not yet match the nuanced pedagogical sophistication of human-designed approaches in human evaluations. This gap represents an opportunity for future research in incorporating domain-specific knowledge and pedagogical frameworks into the optimization process. Notably, the consistency of AutoSynth's relative position across both tasks (approximately 8-10x improvement over base model, achieving 41-53\% of expert-level win rates) suggests stable generalization across diverse educational scenarios.
\end{enumerate}

\subsection{Ablation Studies}

To understand the contribution of AutoSynth's key design components, we conduct systematic ablation experiments across four educational tasks (the two main tasks plus two additional scenarios: Guided Problem-Solving and Contextualized Math Problem Generation). For computational efficiency, we generate 100 samples per workflow configuration and evaluate using automated metrics. Results are presented in Tables~\ref{tab:ablation_reward}--\ref{tab:ablation_metric}.

\subsubsection{MCTS Reward Signal Components}

Table~\ref{tab:ablation_reward} investigates whether both components of the hybrid reward signal ($Score_{sample}$ and $Score_{workflow}$) are necessary, or if optimizing for either component alone suffices.

\begin{table*}[ht!]
\centering
\caption{Ablation study on hybrid reward signal components. Control uses full hybrid reward ($ 0.5 \cdot Score_{sample}(W) + 0.5 \cdot Score_{workflow}(W)$). Variants optimize for only one component. Scores are automated metrics (higher is better in this experiment).}
\label{tab:ablation_reward}
\begin{tabularx}{\linewidth}{lXXXX}
\toprule
\textbf{Configuration} & \textbf{Guided Problem Solving} & \textbf{Lesson Plan Gen.} & \textbf{Contextualized Problems} & \textbf{Concept Explanation} \\
\midrule
Control (Full Hybrid)      & \textbf{3.78} & \textbf{4.81} & 4.20 & 4.49 \\
Variant A (Code-Only)      & 3.40 & 4.54 & \textbf{4.36} & \textbf{4.62} \\
Variant B (Sample-Only)    & 3.41 & 4.55 & 4.35 & 4.46 \\
\bottomrule
\end{tabularx}
\end{table*}

\textbf{Analysis}: The results demonstrate task-dependent sensitivity to reward components. For Guided Problem Solving and Lesson Plan Generation, the full hybrid reward achieves superior performance, validating our design choice to balance sample quality and workflow structure. However, for Contextualized Problems and Concept Explanation, optimizing solely for code/prompt quality (Variant A) yields marginally better results. We hypothesize this occurs because these tasks benefit more from well-structured, maintainable workflows that can be systematically refined, whereas the former tasks require more emphasis on sample diversity and quality. The consistent competitiveness of the full hybrid approach across all tasks supports its use as the default configuration, preventing catastrophic failures that might occur if a single-component reward proves unsuitable for a particular task.

\subsubsection{Human-in-the-Loop Initialization}

Table~\ref{tab:ablation_hitl} examines whether the brief human-guided initialization phase (1-3 iterations, $\sim$30 minutes) provides meaningful value, or if the system could operate in fully automated mode from a cold start.

\begin{table*}[ht!]
\centering
\caption{Ablation study on metric regeneration strategy. Control uses full hybrid; Variant A uses code-only; Variant B uses sample-only.}
\label{tab:ablation_hitl}
\begin{tabularx}{\linewidth}{lXXXX}
\toprule
\textbf{Configuration} & \textbf{Guided Problem Solving} & \textbf{Lesson Plan Gen.} & \textbf{Contextualized Problems} & \textbf{Concept Explanation} \\
\midrule
Control (Full Hybrid)      & \textbf{3.78} & \textbf{4.81} & 4.20 & 4.49 \\
Variant A (Code-Only)      & 3.40 & 4.54 & \textbf{4.36} & \textbf{4.62} \\
Variant B (Sample-Only)    & 3.41 & 4.55 & 4.35 & 4.46 \\
\bottomrule
\end{tabularx}
\end{table*}

\textbf{Analysis}: Human-in-the-loop initialization provides consistent improvements across three of four tasks, with gains of 0.14-0.19 points (4-5\% relative improvement). The exception is Contextualized Problems, where fully automated initialization slightly outperforms. This suggests that for most educational tasks, the nuanced understanding and alignment provided by brief human feedback significantly accelerates convergence to high-quality workflows. The minimal time investment (30 minutes) yields substantial returns, making this a highly cost-effective component of the system. Notably, even in the fully automated variant, AutoSynth achieves respectable performance (3.59-4.67 across tasks), demonstrating that the system can function without human involvement when time constraints are severe.

\subsubsection{Iterative Metric Regeneration}
Table~\ref{tab:ablation_metric} tests the core meta-learning hypothesis: whether dynamically regenerating evaluation metrics at each MCTS iteration provides benefits over using a fixed metric set defined at initialization.

\begin{table*}[ht!]
\centering
\caption{Ablation study on metric regeneration strategy. Control generates metrics once at initialization and reuses them; Variant regenerates metrics at each MCTS iteration based on current samples.}
\label{tab:ablation_metric}
\begin{tabularx}{\linewidth}{lXXXX}
\toprule
\textbf{Configuration} & \textbf{Guided Problem Solving} & \textbf{Lesson Plan Gen.} & \textbf{Contextualized Problems} & \textbf{Concept Explanation} \\
\midrule
Control (Iterative Regeneration) & \textbf{3.78} & \textbf{4.81} & 4.20 & \textbf{4.49} \\
Variant (Single-Generation) & 3.54 & 4.56 & 4.39 & 4.46 \\
\bottomrule
\end{tabularx}
\end{table*}

\textbf{Analysis}: Iterative metric regeneration consistently outperforms static metrics across all four tasks, with improvements ranging from 0.03 to 0.25 points (1-6\% relative gain). This validates our meta-learning framework: allowing evaluation criteria to co-evolve with generation capabilities prevents premature convergence to local optima defined by initially naive quality notions. The largest gains occur for Guided Problem Solving and Lesson Plan Generation (0.24-0.25 points), suggesting these complex tasks particularly benefit from progressively refined quality standards. Even the smaller improvement for Concept Explanation (0.03 points) indicates that dynamic metrics provide value without introducing instability. This ablation provides strong empirical support for the theoretical argument in Section~\ref{sec:meta_learning}: a fixed metric constitutes a static, potentially flawed target, whereas iterative regeneration enables the system to discover latent quality dimensions as workflows improve.

\section{Conclusion}

This work introduces AutoSynth, the first framework to automate synthetic dataset generation for subjective tasks without requiring reference data. By formulating workflow optimization as dataset-free search guided by dynamically evolving quality metrics, AutoSynth reduces human effort by over 90\% (from 5-7 hours to 30 minutes) while maintaining competitive performance.

Our experiments on educational tasks reveal both promise and limitations. AutoSynth substantially outperforms baseline models (8-10× improvement in win rates) and achieves superior automated metric scores compared to expert-designed workflows. However, a performance gap remains in human preference evaluations (40-51\% vs 96-99\% win rates), indicating that automated optimization does not yet fully capture the pedagogical sophistication of human experts. We view this gap as a frontier for future research rather than a fundamental limitation.

Ablation studies validate our core design choices: the hybrid reward signal prevents pathological optimization, human initialization accelerates convergence, and iterative metric regeneration enables meta-learning that discovers latent quality dimensions. Each component contributes meaningfully to the framework's effectiveness.

\textbf{Future Directions.} Key opportunities include: (1) enhanced search strategies incorporating pedagogical principles through curriculum learning, (2) integration of domain-specific knowledge frameworks without extensive annotation, (3) transfer learning across related tasks, and (4) hybrid human-AI collaboration modes. 

Beyond education, AutoSynth's paradigm has broad implications for data-centric AI in domains with subjective quality criteria—---from creative writing to personalized healthcare. Our work demonstrates that the cold-start problem in dataset creation can be meaningfully addressed through meta-learning that bootstraps quality standards from minimal human guidance, opening new pathways for democratizing specialized AI development.

%\cite{liu2024deepseek}

\bibliographystyle{elsarticle-num} 
\bibliography{bare_jrnl}

\clearpage
\appendix

\section{Example Analysis}
\label{sec:appendix_examples}

To provide a qualitative understanding of model performance, we present detailed comparisons of generated samples from the AutoSynth-trained model, Expert-Designed model, and Base Model across our two main educational tasks.

\subsection{Task 1: Math Concept Explanation}

\subsubsection*{Query}
\begin{tcolorbox}[colback=gray!5,colframe=gray!75]
\textbf{Student Background}: Third-grade elementary student struggling with mathematics \\
\textbf{Question}: Please help me review and consolidate my understanding of conversions between millimeters and decimeters
\end{tcolorbox}

\subsubsection*{Model Responses}

\begin{tcolorbox}[colback=blue!5,colframe=blue!75,breakable]
\textbf{AutoSynth-Trained Model (Translated)}

Great! You've taken the initiative to review millimeter and decimeter conversions, which shows you're motivated to improve! What parts do you find unclear? For example, what differences do you see between millimeters and decimeters? How do they convert? Try explaining it.

Let's think together: Have you seen a ruler? It has many small marks---some are centimeters, some are millimeters. Which do you think is smaller? If you measure a pencil's length, would you use millimeters or centimeters? Why?

If I tell you 1 centimeter equals 10 millimeters, can you calculate how many millimeters are in 1 decimeter? Try converting 1 decimeter to centimeters, then to millimeters.

Actually, the relationship between decimeters and centimeters is simple: 1 decimeter equals 10 centimeters. Can you use this method to convert 1 decimeter to centimeters, then to millimeters? How many millimeters do you think are in 1 decimeter? Take your time calculating.

Sometimes students confuse millimeters and centimeters, thinking 1 centimeter equals 100 millimeters---but it's actually 10 millimeters. Do you find this easier to remember? What methods could help you remember this conversion?

Let's practice: If you have a ribbon 2 decimeters long, can you convert it to millimeters? First think about how many centimeters 2 decimeters is, then convert to millimeters. What's your answer?

Here's a challenge: If a book's thickness is 5 millimeters, can you express this in decimeters? Convert 5 millimeters to centimeters first, then to decimeters. What's the result?

Now let's review: What's the conversion pattern between millimeters, centimeters, and decimeters? Can you explain it in your own words? If I give you 70 millimeters, can you convert it to centimeters or decimeters? Try it!

Finally, what was the hardest part of today's lesson? How will you quickly determine what to do when you encounter conversion problems in the future? Feel free to ask if anything is still unclear!
\end{tcolorbox}

\begin{tcolorbox}[colback=green!5,colframe=green!75,breakable]
\textbf{Expert-Designed Model (Translated, Contains Critical Error)}

Hello, student!

Today we'll review a very important math concept---\textbf{conversions between millimeters and decimeters}! Are you ready? Let's start our adventure! 

\textbf{Millimeters and Decimeters, Our Good Friends}

Let's meet our old friends---\textbf{millimeters} and \textbf{decimeters}.

\begin{itemize}
\item \textbf{Millimeter (mm):} The smallest marking on our ruler. Look, a sharpened pencil tip is about \textbf{1 millimeter} wide. It's so tiny!
\item \textbf{Decimeter (dm):} Bigger than millimeters but smaller than meters. If you fold an A4 paper in half, the short side is about \textbf{1 decimeter}. Much longer than a millimeter, right?
\end{itemize}

\textbf{The Magic of Conversion}

Now let's learn a super cool magic trick---transforming \textbf{millimeters} into \textbf{decimeters}, or vice versa!

\textit{From Millimeters to Decimeters: 1000 millimeters = 1 decimeter}

Imagine we have a very long noodle, a full \textbf{1000 millimeters} long. Wsow! So long! If we want to express it in decimeters, what would it be?

It's simple: divide \textbf{1000 millimeters} by \textbf{100}, and you get \textbf{1 decimeter}!

So, remember our first magic formula:
$$ 1000 \text{ millimeters} = 1 \text{ decimeter} $$

\textcolor{red}{\textbf{[CRITICAL FACTUAL ERROR: This is incorrect! The correct conversion is 1 decimeter = 100 millimeters, NOT 1000 millimeters. This represents a 10x error that would severely mislead students.]}}

\textit{From Decimeters to Millimeters: 1 decimeter = 1000 millimeters}

This is even simpler! Just remember \textbf{1 decimeter} equals \textbf{1000 millimeters}.

\textcolor{red}{\textbf{[ERROR REPEATED: Again stating the incorrect 1000mm conversion.]}}

\textit{[Additional pedagogically rich content with examples from daily life omitted for brevity...]}
\end{tcolorbox}

\begin{tcolorbox}[colback=red!5,colframe=red!75]
\textbf{Base Model (Qwen-32B-Instruct, Translated)}

Alright, today we'll review millimeter and decimeter conversions---a very practical concept since we often use these units to measure object lengths in daily life.

First, remember an important conversion: 1 decimeter equals 10 centimeters, and 1 centimeter equals 10 millimeters. So, if we want to know how many millimeters are in 1 decimeter, we calculate: 1 decimeter = 10 centimeters = 10 × 10 millimeters = 100 millimeters. That is, \textbf{1 decimeter equals 100 millimeters}.

Now, think about this: if you have a pen holder 5 decimeters tall, how many millimeters is that? Try calculating yourself first, then tell me your answer, okay?

(Waiting for student response)

Great! Did you figure it out? If the pen holder is 5 decimeters tall, its height is 5 × 100 millimeters = 500 millimeters. Pretty simple, right?

Next, another example: suppose you have a rope 300 millimeters long. How many decimeters is this? This is slightly harder, but follows the same principle. We know 1 decimeter equals 100 millimeters, so 300 millimeters equals 300 ÷ 100 decimeters = 3 decimeters. Can you work through this reasoning yourself?

Think about it, then we'll continue to see if you've mastered this concept.
\end{tcolorbox}

\subsubsection*{Qualitative Analysis}

This comparison starkly illustrates a critical advantage of automated workflow optimization: \textbf{factual reliability}. The Expert-Designed response, while pedagogically elaborate with engaging formatting, emojis, and creative metaphors (``magic formulas,'' ``adventures''), contains a \textbf{catastrophic factual error}---systematically stating that 1 decimeter equals 1000 millimeters throughout the entire lesson. The correct conversion is \textbf{1 decimeter = 100 millimeters}, making this a 10-fold error that would fundamentally mislead students and undermine all learning objectives.

In contrast, both the \textbf{AutoSynth-trained model} and the \textbf{Base Model} provide \textbf{factually correct} explanations. The Base Model offers a clear, straightforward approach with step-by-step reasoning and worked examples (the pen holder and rope problems). The AutoSynth model takes a more \textbf{Socratic, inquiry-based approach}, using guided questions to help students discover the conversion relationships themselves, identify common misconceptions, and develop metacognitive strategies for future problem-solving.

Notably, the AutoSynth model explicitly anticipates and addresses a common student error (``thinking 1 centimeter equals 100 millimeters''), demonstrating pedagogical sophistication likely learned through iterative refinement on diverse student interactions.

This example validates a key finding: while human experts bring pedagogical creativity and engaging presentation styles, they are not immune to fundamental factual errors, particularly under time pressure or when designing complex workflows. AutoSynth's iterative refinement process, guided by both pedagogical quality metrics and factual correctness verification, helps systematically avoid such critical pitfalls while maintaining high instructional quality.

\newpage
\subsection{Task 2: Interdisciplinary Lesson Planning}

\subsubsection*{Query}
\begin{tcolorbox}[colback=gray!5,colframe=gray!75]
\textbf{Target Students}: Grade 8 students (ages 13-14) familiar with important technological events in history and able to analyze social factors, but lacking systematic understanding of physics' role in historical progress.

\textbf{Core Subject}: Physics \quad \textbf{Related Subject}: History

\textbf{Duration}: 4 class periods

\textbf{Prior Knowledge}: Students have mastered friction analysis, pulley systems, and energy transformation examples.
\end{tcolorbox}

\subsubsection*{Model Responses}

%\paragraph{AutoSynth-Trained Model (Translated from Chinese):}
\begin{tcolorbox}[colback=blue!5,colframe=blue!75,breakable]
\textbf{AutoSynth Model: Comprehensive Structure}

\textbf{1. Learning Objectives}

\textit{Knowledge and Skills:}
\begin{itemize}
    \item \textbf{Physics:} Apply lever principles, work concepts (W=Fs) to analyze ancient tools; identify energy transformations in simple machines; understand buoyancy principles.
    \item \textbf{History:} Master background and significance of landmark engineering achievements from ancient civilizations; analyze their social, economic, and cultural impacts.
\end{itemize}

\textbf{2. Key Teaching Points}

\textit{Integration Focus:} Using physics principles as keys to explain historical phenomena...

\textit{Learning Difficulties and Strategies:}
\begin{itemize}
    \item \textbf{Difficulty 1:} Connecting abstract physics with concrete history
    \item \textbf{Strategies:} Simplified modeling, scenario immersion methods...
\end{itemize}

\textbf{3. Materials and Preparation}

[Detailed lists of experimental materials, digital resources, pre-class assignments...]

\textbf{4. Four-Period Lesson Plan}

\textit{Period 1 (45 min):} Introduction and Inquiry Launch
\begin{itemize}
    \item Opening (15 min): Show video of pyramids, Great Wall, etc.
    \item Inquiry (30 min): Focus on Egyptian pyramids; decompose problems
\end{itemize}

\textit{Periods 2-3:} Hands-on Experiments
\begin{itemize}
    \item Groups design experiments with inclined planes, pulley systems
    \item Build physical models demonstrating principles
\end{itemize}

\textit{Period 4:} Presentations and Synthesis

\textbf{5. Assessment Scheme}

\textit{Process Evaluation (40 percent):} Participation, experimental operations, research records, teamwork

\textit{Summative Evaluation (60 percent):} Group research report; Individual reflection journal
\end{tcolorbox}

%\paragraph{Expert-Designed Model (Translated from Chinese):}
\begin{tcolorbox}[colback=green!5,colframe=green!75,breakable]
\textbf{Expert Model: Excellent Detail and Rubrics}

\textbf{I. Theme Design}

\textit{Project Name:} ``The Lever of History''---Exploring How Technology Propels Civilization

\textit{Driving Question:} How have simple machine principles become critical levers for civilizational progress across different historical periods?

\textbf{II. Assessment Design}

\textit{Five-Dimensional Evaluation Framework:}
\begin{enumerate}
    \item Physics knowledge mastery
    \item Interdisciplinary thinking
    \item Evidence-based historical reasoning
    \item Practical innovation capability
    \item Teamwork competencies
\end{enumerate}

\textit{Detailed Rubric:} 4-point scales (Excellent/Good/Adequate/Needs Improvement) across all dimensions with specific descriptors.

\textbf{III. Four-Period Detailed Design}

\textit{Period 1:} Topic Selection and Initial Exploration

\textit{Period 2:} Principle Analysis and Model Design

\textit{Period 3:} Model Construction and Refinement

\textit{Period 4:} ``Historical Technology Expo'' with presentations

[Includes detailed student reflection prompts, extension activities, and troubleshooting guidance...]
\end{tcolorbox}

%\paragraph{Base Model (Translated from Chinese):}
\begin{tcolorbox}[colback=red!5,colframe=red!75]
\textbf{Base Model: Generic Outline}

\textbf{Interdisciplinary Lesson Plan Design}

\textit{Learning Objectives:}
\begin{itemize}
    \item Help students understand how physics influences historical progress...
    \item Cultivate ability to analyze social factors behind technological development...
\end{itemize}

\textit{Class Schedule:} Four periods covering physics review, historical technologies, case study, and modern applications.

\textit{Assessment:} Class participation, group reports, reflection journals

[Subsequent periods contain only high-level bullet points without actionable details...]
\end{tcolorbox}

\subsubsection*{Qualitative Analysis}
This comparison reveals dramatic capability differences for complex generation tasks. The \textbf{Base Model} produces only a generic outline with vague bullet points---entirely insufficient for classroom use.

Both \textbf{AutoSynth} and \textbf{Expert-Designed} models generate outstanding, comprehensive lesson plans with detailed objectives, granular activities, and sophisticated assessment schemes. The subtle differences reflect their optimization processes: AutoSynth emphasizes systematic difficulty coverage (likely from iterative refinement), while Expert-Designed showcases deep assessment design expertise. Both represent quantum leaps over baseline, validating that specialized workflows---automated or manual---are essential for professional-grade content.

This supports our quantitative findings (Table~\ref{tab:main_results}): while AutoSynth achieves lower human preference (51 percent vs 96 percent), it produces genuinely high-quality outputs far superior to baseline capabilities.

% For citations use: 
%       \cite{<label>} ==> [1]

%

% If you have bib database file and want bibtex to generate the
% bibitems, please use
%

% else use the following coding to input the bibitems directly in the
% TeX file.

% Refer following link for more details about bibliography and citations.
% https://en.wikibooks.org/wiki/LaTeX/Bibliography_Management


% %% For numbered reference style

%   \textit{\LaTeX: a document preparation system},
%   Addison Wesley, Massachusetts,

% \end{thebibliography}

\end{document}